\documentclass[journal]{journal}
\usepackage{times}
\usepackage{epsfig}
\usepackage{graphicx}
\usepackage{amsmath}
\usepackage{amssymb}
\usepackage{mathrsfs}
\usepackage{CJK}
\usepackage{multirow}
\usepackage{wrapfig}

\ifCLASSINFOpdf
\else
\fi
\begin{document}
%
\title{Hierarchical Image Classification with A Literally Toy Dataset}
\markboth{Journal of \LaTeX\ Class Files,~Vol.~6, No.~1, January~2007}%
{Shell \MakeLowercase{\textit{et al.}}: Bare Demo of IEEEtran.cls for Journals}
%



\maketitle
\thispagestyle{empty}

\begin{abstract}
Unsupervised domain adaptation (UDA) in image classification remains a big challenge. In existing UDA image dataset, classes are usually organized in a flattened way, where a plain classifier can be trained. Yet in some scenarios, the flat categories originate from some base classes. For example, \emph{buggies} belong to the class \emph{bird}. 
We define the classification task where classes have characteristics above and the flat classes and the base classes are organized hierarchically as hierarchical image classification. 
Intuitively, leveraging such hierarchical structure will benefit hierarchical image classification, \emph{e.g.,} two easily confusing classes may belong to entirely different base classes. 
In this paper, we improve the performance of classification by fusing features learned from a hierarchy of labels. Specifically, we train feature extractors supervised by hierarchical labels and with UDA technology, which will output multiple features for an input image. The features are subsequently concatenated to predict the finest-grained class. 
This study is conducted with a new dataset named Lego-15. Consisting of synthetic images and real images of the Lego bricks, the Lego-15 dataset contains 15 classes of bricks. Each class originates from a coarse-level label and a middle-level label. For example, class \emph{``85080"} is associated with \emph{bricks} (coarse) and \emph{bricks round} (middle). In this dataset, we demonstrate that our method brings about consistent improvement over the baseline in UDA in hierarchical image classification. Extensive ablation and variant studies provide insights into the new dataset and the investigated algorithm.
\end{abstract}


%
\IEEEpeerreviewmaketitle

\section{Introduction}
\begin{figure}[t]
\begin{center}
   \includegraphics[width=0.8\linewidth]{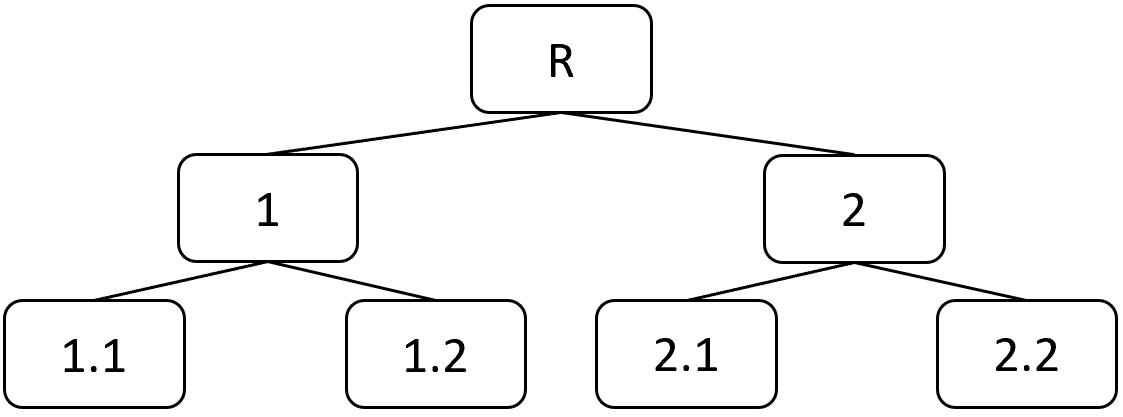}
\end{center}
\caption{Tree structured class hierarchy. Every node represents a sub-class of the class that its parent-node represents.}
\label{fig:class_hierarchy_example}
\end{figure}

%
%
%
%
\IEEEPARstart{U}{DA} in image classification remains a big challenge. In existing UDA dataset, classes are usually organized in a flattened way, like in most image datasets.
Yet in some multiple classes classification scenarios, the flat categories originate from some base classes, and all classes are organized in a hierarchical way, as shown in Fig.~\ref{fig:class_hierarchy_example}. Intuitively, if classes in a UDA dataset have such a hierarchical structure, leveraging such hierarchical structure will benefit classification on the basis of UDA technology.
To make good use of the hierarchical relationship among classes, namely class hierarchy, it is natural to use the hierarchical classification method. Hierarchical classification is a classification approach that can decompose the original classification task whose classes are organized hierarchically into multiple sub-tasks with a smaller scale, therefore, reduce the difficulty of original task. In the real world, many classification problems can be regarded as hierarchical classification problems, such as text classification~\cite{DBLP:conf/cidm/MayneP09}, protein function prediction~\cite{DBLP:journals/pr/TrigueroV16}, image annotation~\cite{DBLP:journals/pr/DimitrovskiKLD11}, etc. However, hierarchical classification suffers from the problem of error propagation, which causes the performance of a certain level will directly affect the performance of its next level, and then affect the final performance.

\begin{figure}
\begin{center}
\includegraphics[width=0.8\linewidth]{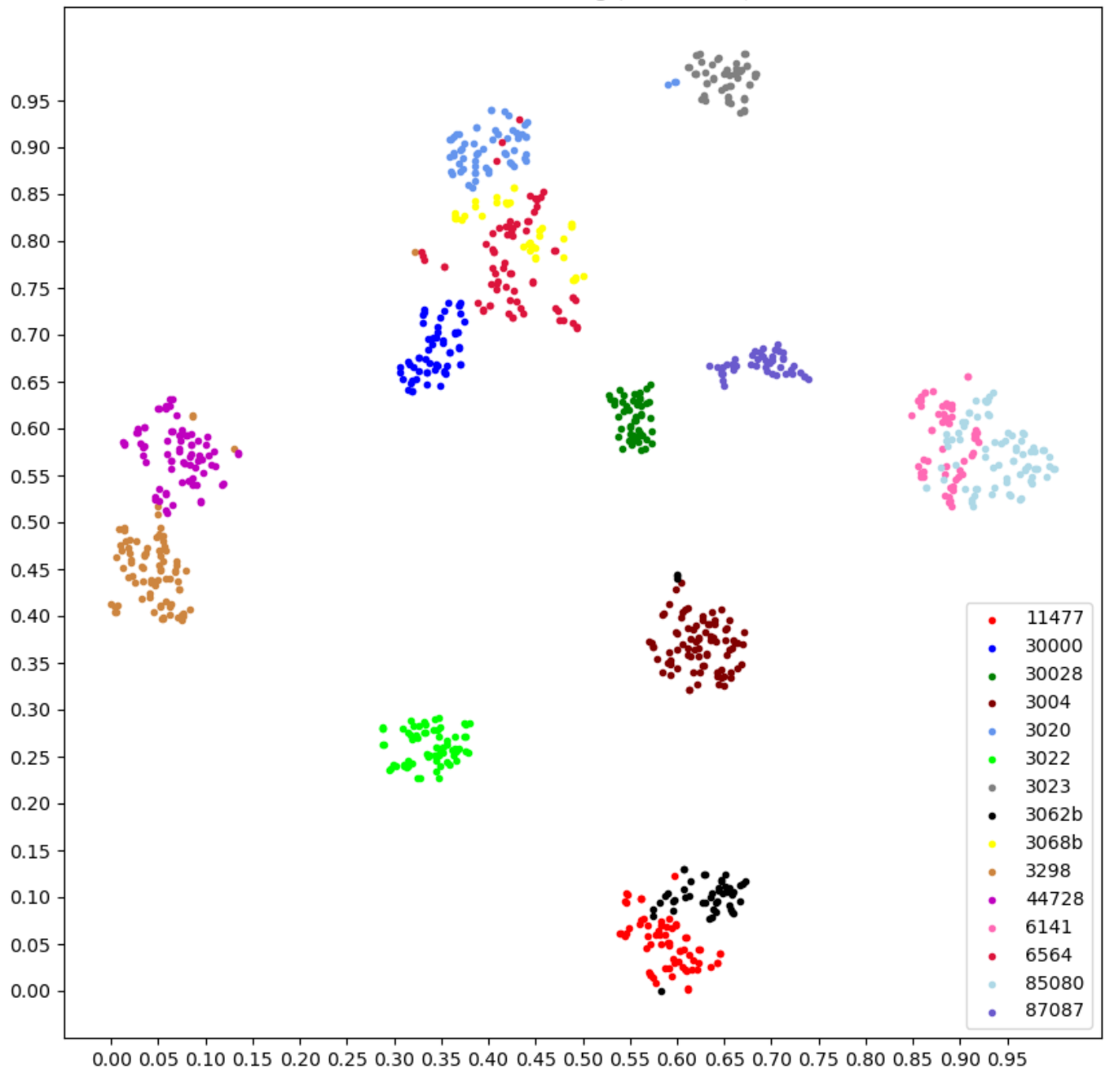}
\end{center}
\caption{T-SNE visualization of features of 15 classes of Lego brick images.}
\label{fig:tSNE_features3_labels_level3_ours_normal_test}
\end{figure}

In many cases, categories are confusing because of their similar appearances. Their features are usually close or even mixed together. For our Lego-15 dataset which will be introduced later and shown in Fig.~\ref{fig:difference_between_src_and_tgt_imgs}, class \emph{``85080"} and class \emph{``6141"} are confusing because of their same color. As shown in Fig.~\ref{fig:tSNE_features3_labels_level3_ours_normal_test}, their features are mixed, which makes it easy for the classifier to wrongly classify features of one class as the other class. If such features are used for classification, the performance will be seriously affected.

\begin{figure}
\begin{tabular}{cc}
\includegraphics[width=0.45\linewidth]{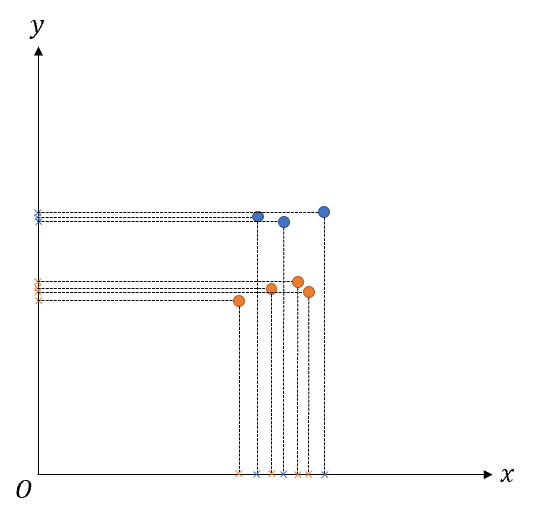} & \includegraphics[width=0.45\linewidth]{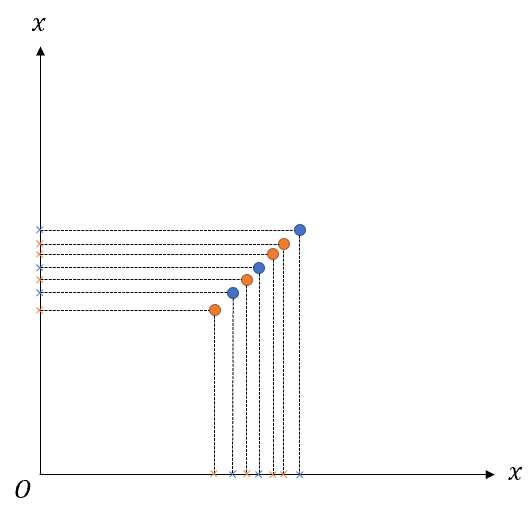}\\
(a) & (b)\\
\end{tabular}
\caption{We use an example where new features are formed by concatenating two hierarchical features to interpret our idea. Blue points and orange points represent new features of two classes. A point's projection on an axe represent its hierarchical feature at a certain granularity. (a) shows that we can discriminate two easily confusing classes if their certain granularity of hierarchical features are discriminative, even though their certain granularity of hierarchical features are highly mixed. (b) shows that concatenating a feature with itself does not increase the overall discriminability.}
\label{fig:rise_dimension_example}
\end{figure}

In order to take the advantage of class hierarchy and overcome the disadvantages of hierarchical classification approach, in this paper, we propose a hierarchical feature fusion classification framework. This framework consists of multiple feature extractors and a classifier. Feature extractors are supervised by a hierarchy of labels, and output multiple features for an input image. The features are subsequently concatenated and then input into the classifier to predict the finest-grained class. Our intuition is that if feature extractors are supervised by hierarchical labels and a sample is represented by a feature formed by concatenating multiple hierarchical features of that sample, two easily confusing classes may be easier to separate, especially when they belong to different base classes. Hence, we concatenate hierarchical features together to get discriminative features, as shown in Fig.~\ref{fig:rise_dimension_example}.

Among the many applications of image classification, an interesting one is Lego image classification. If you look closely, you can find that there exists a hierarchical relationship among many Lego brick classes. For example, classes \emph{``85080"} and \emph{``3062b"} belong to class \emph{bricks round}, while \emph{bricks round} and \emph{bricks special} belong to class \emph{bricks}. 

In order to prove the effectiveness of our method, we introduce a Lego-15 dataset and conduct extensive experiments on it. As we expect, the experimental results sufficiently prove that our method brings a significant performance boost over the baseline. The main contributions of this paper can be summarized as follows:
\begin{itemize}
\item We propose a hierarchical feature fusion method by fusing features learned from a hierarchy of labels for classification tasks that have a class hierarchy.
\item We present a new dataset named Lego-15, which consists of more than 1000 real images and 3000 synthetic images and both have 15 classes of Lego bricks. Each image is carefully annotated and has three labels of different semantic levels. This dataset can be used for further research of UDA in classification tasks that have a class hierarchy.
\item We apply the hierarchical feature fusion method on the Lego-15 dataset and conduct extensive experiments to fully prove the effectiveness of our method.
\end{itemize}

\section{Related Work}
\subsection{Feature Fusion}
In many computer vision tasks (object detection and image segmentation), feature fusion is an important way to improve performance. These features are usually multi-scale, covering the convolution results of different convolution layers in the deep network. In general, feature maps from shallow layers have higher resolution and weaker semantics, while feature maps from deep layers have lower resolution and stronger semantics. How to fuse the complementary features is the key to improve task performance. In this regard, predecessors have carried out a lot of research.

According to the sequence of fusion and prediction, feature fusion methods can be divided into early fusion ~\cite{DBLP:conf/cvpr/BellZBG16,DBLP:conf/cvpr/KongYCS16} and late fusion~\cite{DBLP:conf/eccv/LiuAESRFB16,DBLP:conf/eccv/CaiFFV16,DBLP:conf/cvpr/LinDGHHB17}. In the early fusion method, multiple features are fused into a new one, which is used to train the predictor. The main ways of fusion are concatenation and element-wise addition, etc. The late fusion is to improve the detection performance by combining the prediction results of different layers. In single shot multibox detector (SSD) ~\cite{DBLP:conf/eccv/LiuAESRFB16}, multi-scale CNN (MS-CNN)~\cite{DBLP:conf/eccv/CaiFFV16}, prediction results from multi-scale features are integrated to improve the detection performance. In FPN~\cite{DBLP:conf/cvpr/LinDGHHB17}, the features from different layers are arranged into a pyramid structure to get multiple predictions, and these predictions are integrated to improve final performance.

\subsection{Deep Metric Learning}
Metric learning studies how to learn a distance function on a specific task, so that the distance function can help the nearest neighbor-based algorithms (KNN~\cite{DBLP:journals/tit/CoverH67}, etc.) achieve better performance. Deep metric learning is a method of metric learning. Its goal is to learn a mapping from the original feature to the low dimensional and dense vector space (called embedding space) so that when using the commonly used distance functions (Euclidean distance, cosine distance, etc.) to calculate the distance between samples in the embedding space, the distance between samples of the same class is closer and distance between samples of different classes is further. Deep metric learning has many successful applications in the computer vision field, such as face recognition~\cite{DBLP:conf/cvpr/WangWZJGZL018}, image retrieval~\cite{DBLP:journals/corr/abs-2103-11781}, person re-identification~\cite{DBLP:journals/tip/YangWT18} and so on.

Loss functions play a very important role in deep metric learning, which can be divided into two types: pair-based loss function and classification-based loss function. Among these loss functions, triplet loss~\cite{DBLP:conf/cvpr/SchroffKP15}, contrastive loss~\cite{DBLP:conf/cvpr/HadsellCL06} etc. are classic and effective loss ones. During recent years, there has been remarkable progress in deep metric learning~\cite{DBLP:conf/cvpr/WangWZJGZL018,DBLP:conf/mm/WangXCY17,DBLP:conf/cvpr/DengGXZ19,DBLP:conf/iclr/0015LDL018}. The loss function used in this paper is triplet loss, which is a pair-based loss function.

\section{Task Formulation}

Our task is UDA in image classification. We are given a source domain $D_s=\{(x_i^s,y^s_{i,1},y_{i,2}^s,y_{i,3}^s)\}_{i=0}^{N_s}$ ($y^s_{i,j}\in Y_j, j=1, 2, 3$) of $N_s$ labeled samples, where $y_{i,1}^s, y_{i,2}^s, y_{i,3}^s$ represent labels of sample $x_i^s$ in three levels respectively and $y_{i,3}^s$ is the finest-grain label, and given a target domain $D_t=\{x_i^t,y^t_{i,1},y_{i,2}^t,y_{i,3}^t)\}_{i=0}^{N_t}$ of $N_t$ samples, whose labels are only available when validating or testing. The source domain and target domain obey joint probability distributions $P(X^s, Y_3^s)$ and $Q(X^t, Y_3^t)$ respectively, and $P \neq Q$.  We  assume  source  domain  and  target domain share the same categories. We train our model in source domain and test it in target domain.

\begin{figure*}
\begin{center}

\includegraphics[width=0.9\linewidth]{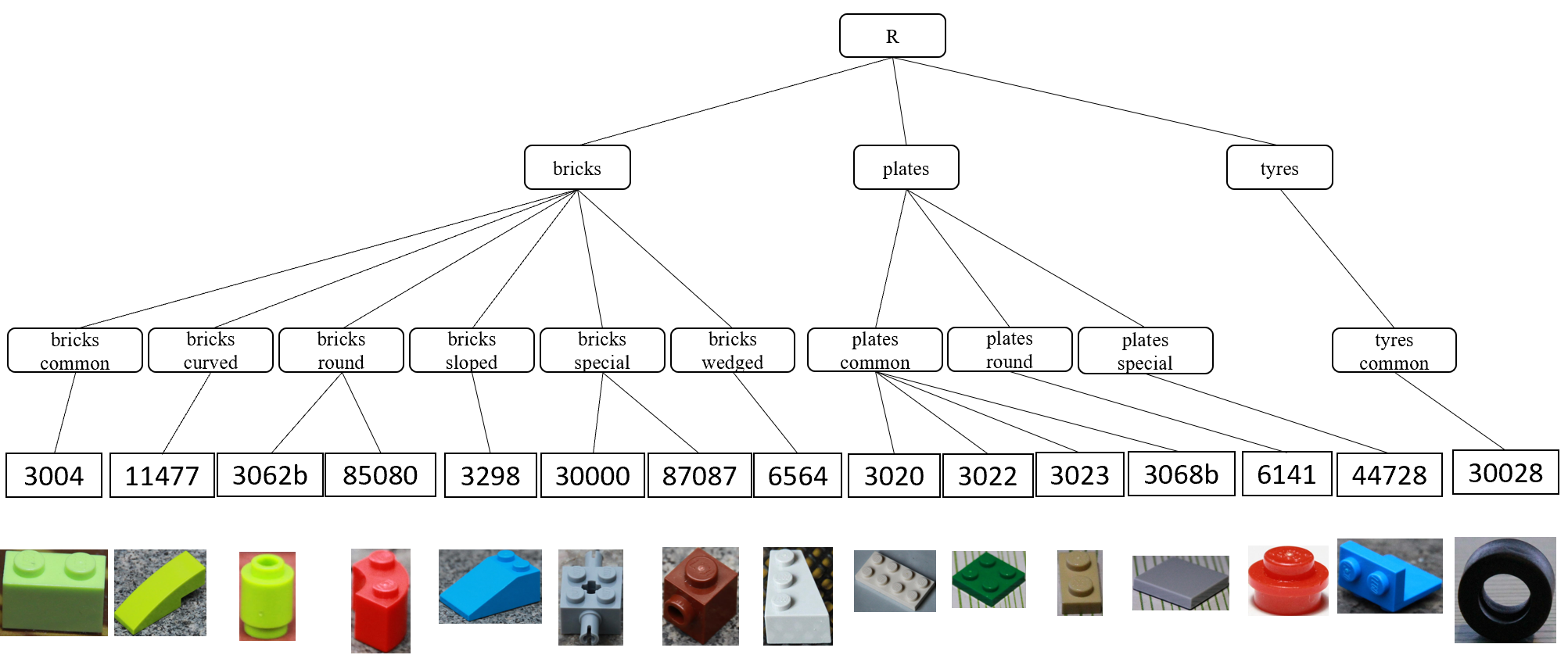}
\end{center}
    \caption{Class hierarchy we predefine.}
    \label{fig:predefine_class_hierarchy}
\end{figure*}

\section{Proposed Framework}
In this section, we present an overview of the proposed framework first and then describe the training of the model.

\subsection{System Overview}
\begin{figure}
\begin{tabular}{c}
\includegraphics[width=\linewidth]{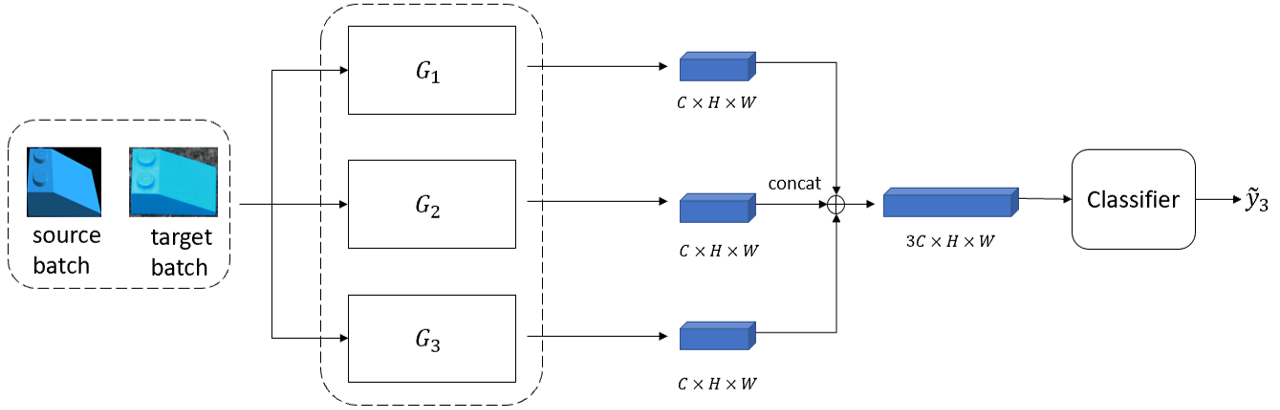} \\
(a)\\
\includegraphics[width=\linewidth]{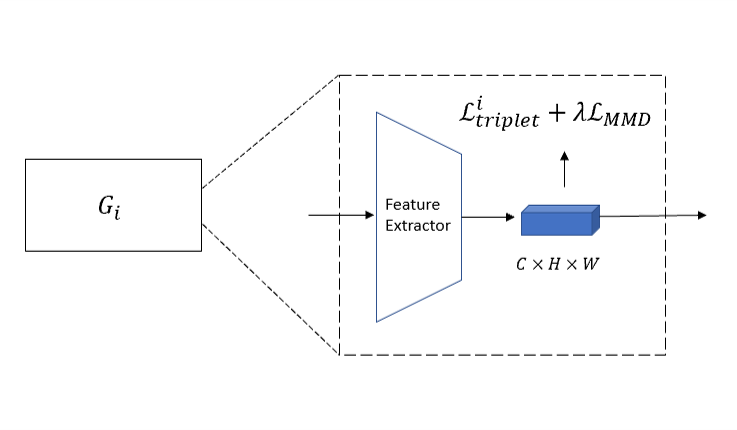}\\
(b)\\
\end{tabular}
\caption{The architecture of our system. For clarity, we divide the whole figure into (a) data flow diagram and (b) specific training method. $\tilde{y}_3$ means predictions for source images or target images. $G_i$ means feature extractor trained with $i$th level labels, and $\mathcal{L}_{MMD}$ is calculated by source batch and target batch inputted system simultaneously. }
\label{fig:network_framework}
\end{figure}

\textbf{Architecture}
The architecture of our system is shown in Fig.~\ref{fig:network_framework}. In our framework, there are 4 components: three hierarchical feature extractors $G_1,G_2,G_3$, and the final classifier $C$. Firstly, three feature extractors extract three features of an image. Each hierarchical feature extractor is trained with the labels of a semantic level so that it can pay attention to features in this level. Then, after the three hierarchical features are obtained, these features will be concatenated into a new feature, so the new feature contains information of three semantic levels. Finally, the final classifier takes the new feature as the input and gives a fine-level class prediction of it.

\textbf{Objective Functions}
Feature extractors are trained with triplet loss and multi-kernel maximum mean discrepancy loss function~\cite{DBLP:conf/icml/LongC0J15}, where the semantic levels of labels used by each feature extractor to calculate triplet loss are different. So the objective function of feature extractor is
\begin{equation}
    \min (\mathcal{L}_{triplet}+\lambda\mathcal{L}_{MMD}).
\end{equation}
Final classifier is trained with cross entropy loss calculated by fine-level labels, so the objective function of classifier is 
\begin{equation}
    \min \mathcal{L}_{CE}.
\end{equation}

\subsection{Hierarchical Feature Extraction}
In the training process, all source images and part of target images are used, but only source images are labeled. In order to extract three different features $f_1,f_2,f_3$ which represent the same sample, a source image is sent to $G_1,G_2$ and $G_3$ as their input. Then, three hierarchical feature extractors output three hierarchical features, representing an image’s three features in three semantic scales respectively. To make features in each level discriminative, $j$th hierarchical feature extractor is trained with triplet loss: 
\begin{equation}  
	\mathcal{L}_{triplet}^j=\frac{1}{N}\sum_{(x^a, x^p, x^n)}\max(\|f^a_j-f^p_j\|-\|f^a_j-f^n_j\|+\alpha, 0),
\end{equation}
where $N$ is the mini-batch size, $(x_a,x_p,x_n)$ is a hard triplet, $f_j^a$, $f_j^p$, $f_j^n$ is the features of $j$th level of the anchor image (j=1,2,3), the positive image and the negative image, separately. $\alpha$ is a margin that is enforced between positive and negative  pairs. Under different semantic scales, three samples do not always form a triplet as they used to do in a certain semantic scale.  For example, $x_1,x_2$ and $x_3$ form a triplet in fine-level class, but they belong to the same middle-level class, so they can’t form  a triplet in middle-level. Using labels of different levels to minimize the triplet loss function can guide the feature extractors to extract information in different semantic levels. 

To decrease the domain gap between source image features and target image features, an extra multi-kernel maximum mean discrepancy loss function \cite{DBLP:conf/icml/LongC0J15} (MK-MMD, hereinafter referred to as MMD) is also used to optimize the feature extractors. Consequently, the total loss function of a feature extractor is 
\begin{equation}
\mathcal{L}\\_{FE}=\mathcal{L}\\_{triplet}+\lambda\mathcal{L}\\_{MMD},
\end{equation}
where $\lambda$ is trade-off between triplet loss function and MMD loss function.

The three features complement each other from three aspects, enrich the semantic information of the sample, and form a feature set that is more complete to describe the sample.

The prediction given by the final classifier is the final prediction result. Before getting the final prediction, we need to concatenate the three hierarchical features into a new one. Then, we input the new feature into the final classifier to get the final fine-level prediction. The final classifier is trained with cross entropy loss function
\begin{equation}
\mathcal{L}\\_{CE}=E_{x\sim\\D_s}[\ell(y_3,C(cat(f_1,f_2,f_3)))],    
\end{equation}
where $C$ denotes the final classifier, $cat(\cdot)$ denotes concatenation operation, $\ell$ denotes cross entropy loss function.

\section{Experiment}
\subsection{The Lego-15 Dataset}
We introduce a Lego-15 dataset. Lego-15 is a Lego image dataset consisting of 3000 synthetic images and 1688 real images in 15 classes. For synthetic images, there are 200 images in each class. For real images, the number of images ranges from 70 to 150 in each class. Synthetic images are rendered by Unity, and domain randomization~\cite{DBLP:conf/iros/TobinFRSZA17} technique is resorted to change image attributes, e.g. illumination and distance. Real images are taken by a camera in the real environment. Multiple backgrounds are chosen to increase the variety of real images. 

We predefine a class hierarchy for the Lego-15 dataset, as shown in Fig.~\ref{fig:predefine_class_hierarchy}. This class hierarchy is represented by a tree. Ignoring the root node, the class hierarchy contains three levels. The third level contains 15 classes, which we call fine-level classes. The 15 fine-level classes can be divided into 10 middle-level classes in the second level and 10 middle-level classes can be further divided into three coarse-level classes in the first level. Apparently, the higher the level, the higher the semantic level, the less the number of classes. We denote three labels space as $Y_1, Y_2$ and $Y_3$.

\begin{figure}
\begin{center}
    \fbox{{\includegraphics[width=\linewidth]{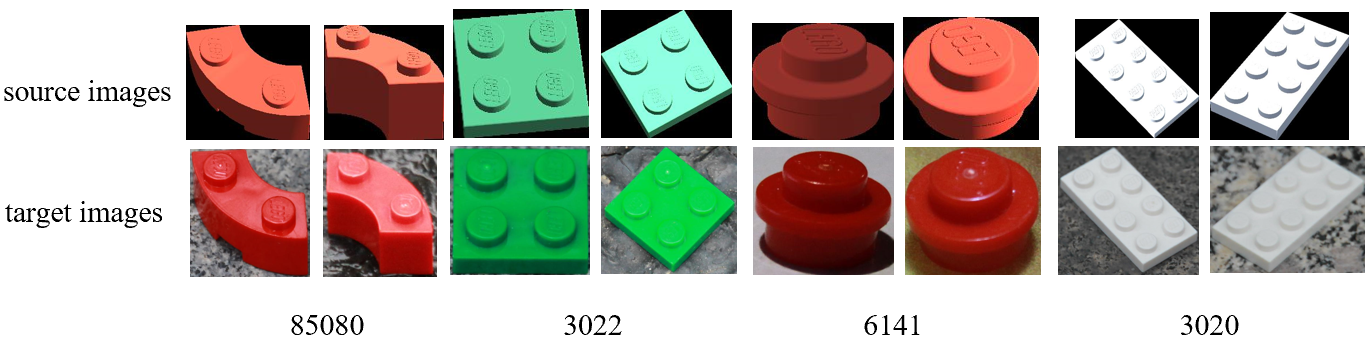}}}
\end{center}

\caption{Difference between source images and target images.}
\label{fig:difference_between_src_and_tgt_imgs}
\end{figure}

The difference between synthetic images and real images can be seen in Fig.~\ref{fig:difference_between_src_and_tgt_imgs}. Obviously, two kinds of images present large domain gap because of the imperfection of the simulator, so we refer to synthetic images as source images, and real images as target images. Source images are all labelled and are denoted as $D_s=\{(x_i^s,y^s_{i,1},y_{i,2}^s,y_{i,3}^s)\}_{i=0}^{N_s}$ ($y^s_{i,j}\in Y_j, j=1, 2, 3$), where only $y_{i,3}^s$ is an image’s original label, $y_{i,1}^s$ and $y_{i,2}^s$ are manually annotated according to the class hierarchy. Target images are denoted as $D_t=\{x_i^t,y^t_{i,1},y_{i,2}^t,y_{i,3}^t)\}_{i=0}^{N_t}$, but their labels are only available when validating or testing. We assume source domain and target domain share the same label space.

We split the whole dataset into the training set, the validation set, and the testing set. The training set contains 3000 source domain images and 750 target domain images, where target images are randomly selected in proportion among 15 fine-level classes. The validation set contains 75 target images in 15 classes, with 5 images in each class. The testing set contains the remaining 863 target images.

\subsection{Experimental Setup}
\textbf{Evaluation protocols of Lego-15.}
To evaluate the overall classification accuracy of a method on the Lego-15 testing set, we set up our evaluation protocol based on top-1 classification accuracy.

\textbf{Baseline structures. }
Like the hierarchical feature fusion framework, the baseline is consists of three feature extractors and one final classifier, whose network weights are initialized the same as those of their counterparts in our method. Different from ours, labels used in the baseline are all fine-level labels that have 15 categories. The purpose of this is to control the network capacity and the feature length used for the final classification of the baseline and of ours to be the same, and only keep labels to be different, so as to better study the impact of hierarchical feature fusion on the final classification performance by comparing the differences of classification performance between them.

\textbf{Implementation details.}
We implement our experiments on widely used Pytorch. We train the model on NVIDIA GeForce GTX 1080 GPU, with 11GB graphic card memory. Without loss of generality, we use ResNet18~\cite{DBLP:conf/cvpr/HeZRS16} as our backbone, that is, we use the remaining network after removing the fully connected layer as the feature extractor. Our approach can also be applied to other deep convolutional neural networks. $C$ is a single-layer fully connected network with input length 1536 (the length of a hierarchical feature is 512, so the length of the fused feature is 1536) and output length 15. We use mini-batch SGD as our optimizer with a learning rate of 0.0001, a momentum of 0.9, and a weight decay of 0.0005 for three feature extractors and the final classifier. The batch size is set as 8. Each network is trained for 60 epochs and tested directly on the target images after the training. We run our whole learning process 3 times with different random seeds and report the average top-1 accuracy.

\begin{table}
    \centering
    \begin{tabular}{lcccc}
        \hline
        Method & $FE_1$ & $FE_2$ & $FE_3$ & Acc \\
        \hline
        baseline & 3 & 3 & 3 & 64.31\% \\
        baseline w. coarse & 1 & 3 & 3 & 66.16\% \\
        baseline w. middle & 3 & 2 & 3 & 65.24\% \\
        ours & 1 & 2 & 3 & 69.41\% \\
        \hline
    \end{tabular}
    \caption{Ablation studies of coarse-level and middle-level labels. $FE_i$ means $i$th feature extractor. Numbers under $FE_i$ mean what semantic level of labels $i$th feature extractor use. We denote coarse level as 1, middle level as 2 and fine level as 3.}
    \label{tab:ablation_study}
\end{table}

\begin{table}
    \centering
    \begin{tabular}{lc}
\hline
Method & Final acc \\
\hline
Baseline($\mathcal{L}_{triplet}$) & 64.31\% \\
Ours($\mathcal{L}_{triplet}$) & 69.41\% \\
\hline
    \end{tabular}
\caption{Top-1 accuracy of different methods on Lego  dataset. }
\label{tab: top_1_acc}
\end{table}

\subsection{Evaluation}
\textbf{Effectiveness over the baseline}
We compare our method with baseline (both are trained with domain adaptation) on the Lego-15 testing set, and the results are reported in Table \ref{tab: top_1_acc}. Our method significantly outperforms the baseline, which proves that our method benefits classification on the basis of UDA technology.

\begin{table}
\begin{center}
\begin{tabular}{lc}
\hline
Method & Final acc \\
\hline
Baseline(ResNet18) & 64.31\% \\
Baseline(GoogLeNet) & 67.32\% \\
Ours(ResNet18) & 69.41\% \\
Ours(GoogLeNet) & 70.34\% \\
\hline
\end{tabular}
\end{center}
\caption{Top-1 accuracies of methods using different backbones on Lego-15 dataset.}
\label{tab:top_1_acc_backbone}
\end{table}

\textbf{Impact of different backbones} We use ResNet18 as our feature extractors by default, and we also use different backbones as feature extractors to study the impact of different backbones on our method and the baseline. For a fair comparison, we use GoogLeNet, whose amount of parameters is close to that of ResNet18, as the comparison target of ResNet18. The top-1 classification accuracies of using GoogLeNet as the backbone are shown in Table \ref{tab:top_1_acc_backbone}. From Table \ref{tab:top_1_acc_backbone}, we can see that GoogLeNet clearly outperforms ResNet18 with a small margin. We assert that this is because GoogLeNet has more network parameters than ResNet18.

\textbf{Necessity of performing domain adaptation.} The top-1 accuracy of the baseline and our method trained with and without domain adaptation is shown in Table \ref{tab:necessity_of_DA}. We clearly observe that performing domain adaptation improves the performance of both the baseline and our method and thus is necessary.

\textbf{Ablation studies of coarse-level and middle-level labels.} We analyze the contribution of the coarse-level labels and the middle-level labels. Table \ref{tab:ablation_study} shows the comparison results. From this table, we can summarize that:

Firstly, the introduction of coarse-level label and middle-level label improves the classification accuracy. Table \ref{tab:ablation_study} shows that method ``baseline w. coarse" and ``baseline w. middle" both outperform method ``baseline" by a clear margin. Our method, namely method after introducing coarse-level label and middle-level label to the baseline, achieves the best performance. Secondly, we can see that the introduction of coarse-level label boosts performance better than the introduction of middle-level label. We argue that this is because concatenation with coarse-level feature increases the discriminability of two classes of features that are close originally more significantly than concatenation with middle-level feature. We will discuss it in Sec. \ref{sec:analysis}.

\begin{table}[]
    \centering
    \begin{tabular}{lc}
        \hline
        Method & Acc \\
        \hline
        baseline w.o. DA & 59.91\% \\
        ours w.o. DA & 57.47\% \\
        \hline
        baseline w. DA & 64.31\% \\
        ours w. DA & 69.41\% \\
        \hline
    \end{tabular}
    \caption{Top-1 accuracy of the baseline and our method w/ and w/o domain adaptation.}
    \label{tab:necessity_of_DA}
\end{table}

\begin{figure}
\centering
\includegraphics[width=\linewidth]{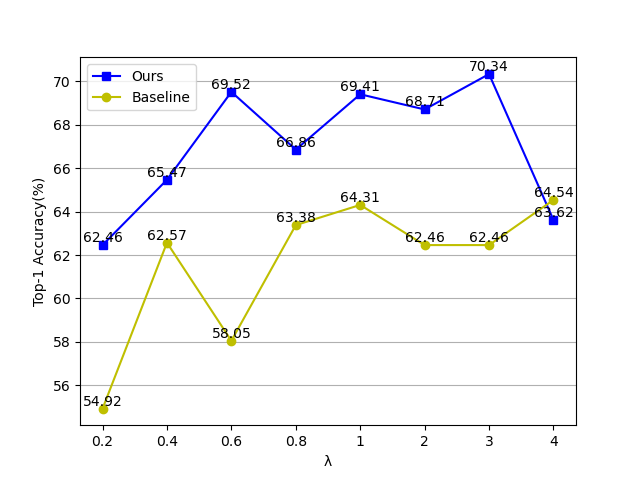}
\caption{The top-1 accuracies of models trained with different $\lambda$ on Lego-15 dataset.}
\label{fig:hyperparameter_analysis_lambda_mmd}
\end{figure}

\textbf{Sensitivity of hyper-parameters.} The hyper-parameter $\lambda$ is used to control the trade-off between $\mathcal{L}_{triplet}$ and $\mathcal{L}_{MMD}$. We set $\lambda=1$
by default, and we also conduct an experiment to investigate its sensitiveness. In the experiment, we vary $\lambda$ from 0.2 to 4 to train different models. The top-1 accuracies of these models on Lego-15 dataset are illustrated in Fig.~\ref{fig:hyperparameter_analysis_lambda_mmd}. It is clear that the top-1 accuracy varies much across a wide range of $\lambda$. But our method outperforms baseline under almost all values of $\lambda$. We can also observe that too high or too low a value of $\lambda$ will both lead to poor performance of ours or the baseline. Properly choosing the $\lambda$ value will make the baseline and our method both achieve good performances.

\begin{table*}
    \centering

        \begin{tabular}{lccccc}
        \hline
        Method & 3068b(6564) & 44728(3298) & 6564(3020) & 85080(6141) & 87087(85080) \\
        \hline
        baseline & 35\%(59\%) & 46\%(53\%) & 73\%(4\%) & 35\%(64\%) & 50\%(38\%) \\
        ours & 74\%(0\%) & 88\%(11\%) & 38\%(15\%) & 96\%(3\%) & 68\%(18\%) \\
        \hline
    \end{tabular}
    \caption{Values of $\mathcal{M}(\cdot, \cdot)$. In the first row, from the second columns to the sixth columns, the class inside brackets is the class as which features of class outside brackets most likely to be wrongly classified. We denote class outside brackets as $C_1$, class inside brackets as $C_2$, every pair of numbers in the second row and the third row means $\mathcal{M}(C_1, C_1)$ ($\mathcal{M}(C_1, C_2)$).   }

    \label{tab:max_percent}
\end{table*}

\section{Analysis}
\label{sec:analysis}
In this section, we investigate the reason for the superiority of our method.

{\bf Why do we concatenate features?} Supposing there are two feature spaces and two classes of images whose features are hard to discriminate from features of different classes in the first feature space but easy to discriminate in the second feature space. If we can concatenate two features to form a new feature, the feature formed by concatenating feature in the first feature space with its counterpart in the second feature space must be easier to discriminate than feature formed by concatenating feature in the first feature space with itself, as illustrated in Fig.~\ref{fig:rise_dimension_example}. The reason is that concatenation itself does not increase the discriminability of new features, but the high discriminability of the original features does. Consequently, we concatenate features to form discriminative new features, on the premise that features are discriminative in at least one feature space.


{\bf Why is our method better than baseline?} The only difference between our method and the baseline is labels used. Our method uses hierarchical labels, while the baseline only uses finest-grained labels. Three features of an image obtained by using our method may be very different but may be almost the same using the baseline. Since two easily confusing fine-level classes probably have discriminative coarse-level features or middle-level features if they belong to different coarse-level classes or middle-level classes, features formed by concatenating three hierarchical features together are likely to be more discriminative than those formed by concatenating only features of fine-level together. Hence, our method is better than the baseline. To validate our explanation, We define $\mathcal{M}(C_1, C_2)$:
\begin{equation}
    \mathcal{M}\\(C_1, C_2) = \frac{\sum_{y^t_{i,3}=C_1}\mathcal{I}\\(\mathcal{H}\\(x_i^t), P_{C_2})}{\sum_{y^t_{i,3}=C_1}1},
\end{equation}
\begin{equation}
    \mathcal{I}(f, P_i)=\left\{
\begin{aligned}
& 1,\ if\ CS(f, P_i) > CS(f, P_j) \ {\forall} j \in Y_3\ (i \ne j) \\
& 0,\ else
\end{aligned}
\right.,
\end{equation}
where $C_1$ and $C_2$ denote two fine-level classes, $\mathcal{H}(x)$ outputs the fused feature of an image $x$, $P_i$ denotes class prototype of class $i$ in fine-level classes, and $CS(\cdot, \cdot)$ denotes cosine similarity. We use $\mathcal{M}(C_1, C_2)$ to measure the possibility that samples belonging to $C_1$ are classified as $C_2$.

Calculations of $\mathcal{M}(\cdot, \cdot)$  are presented in Table \ref{tab:max_percent}. As Table \ref{tab:max_percent} demonstrates, after applying our method, more samples achieve maximum cosine similarity with their correct class prototypes, and fewer samples achieve maximum cosine similarity with prototypes of classes as which they are most likely to be wrongly classified. It evidences that our method makes features of two easily confusing classes more positive than the baseline does, thereby yielding a significant improvement of overall accuracy over the baseline.

\section{Conclusion}
In this paper, we introduce class hierarchy to UDA in image classification, and propose a hierarchical feature fusion method. To prove the effectiveness of our method, we construct a UDA dataset Lego-15 and conduct experiments on it.  
Experimental results demonstrate a noticeable improvement of our method over the baseline on the proposed dataset, which strongly evidences that our method can further improve classification performance on the basis of UDA technology.

\ifCLASSOPTIONcaptionsoff
  \newpage
\fi

\end{document}